%% file: LVM_CCA.tex
\documentclass[peerreview]{IEEEtran}
\newcommand{\mvgauss}[3]{\mathcal{MN}({#1},{#2},{#3})} % this defines a matrix variate gaussian
\newcommand{\gauss}[2]{\mathcal{N}({#1},{#2})} % this defines a gaussian

\usepackage{cite}
\ifCLASSINFOpdf
\else
\fi
\usepackage{amsmath}
\usepackage{amssymb}
\usepackage{algorithm}
\usepackage{tikz}
\usepackage{pgfplots}
\usepackage{booktabs}
\usepgfplotslibrary{units,groupplots}
\pgfplotsset{compat=newest}
\usetikzlibrary{fit,positioning}
\usetikzlibrary{bayesnet}
\usetikzlibrary{plotmarks}

\usepackage{algorithmic}
\DeclareMathOperator*{\argmax}{arg\,max}

\begin{document}
%
% paper title
% Titles are generally capitalized except for words such as a, an, and, as,
% at, but, by, for, in, nor, of, on, or, the, to and up, which are usually
% not capitalized unless they are the first or last word of the title.
% Linebreaks \\ can be used within to get better formatting as desired.
% Do not put math or special symbols in the title.
\title{a Latent Variable Model for Two-Dimensional Canonical Correlation Analysis and its Variational Inference}
%
%
% author names and IEEE memberships
% note positions of commas and nonbreaking spaces ( ~ ) LaTeX will not break
% a structure at a ~ so this keeps an author's name from being broken across
% two lines.
% use \thanks{} to gain access to the first footnote area
% a separate \thanks must be used for each paragraph as LaTeX2e's \thanks
% was not built to handle multiple paragraphs
%

\author{Mehran~Safayani,
        and~Saeid~Momenzadeh
        % <-this % stops a space
        \thanks{M. Safayani is with the Department of Electrical and Computer Engineering, Isfahan University of Technology,
Isfahan 84156-83111, Iran e-mail: (safayani@cc.iut.ac.ir, (corresponding author)).}% <-this % stops a space
        \thanks{S. Momenzadeh is with the Department of Electrical and Computer Engineering, Isfahan University of Technology,
Isfahan 84156-83111, Iran e-mail: (s.momenzadeh@ec.iut.ac.ir).}}
%\thanks{Manuscript received April 19, 2005; revised August 26, 2015.}}

% note the % following the last \IEEEmembership and also \thanks -
% these prevent an unwanted space from occurring between the last author name
% and the end of the author line. i.e., if you had this:
%
% \author{....lastname \thanks{...} \thanks{...} }
%                     ^------------^------------^----Do not want these spaces!
%
% a space would be appended to the last name and could cause every name on that
% line to be shifted left slightly. This is one of those "LaTeX things". For
% instance, "\textbf{A} \textbf{B}" will typeset as "A B" not "AB". To get
% "AB" then you have to do: "\textbf{A}\textbf{B}"
% \thanks is no different in this regard, so shield the last } of each \thanks
% that ends a line with a % and do not let a space in before the next \thanks.
% Spaces after \IEEEmembership other than the last one are OK (and needed) as
% you are supposed to have spaces between the names. For what it is worth,
% this is a minor point as most people would not even notice if the said evil
% space somehow managed to creep in.

% The paper headers
\markboth{Journal of \LaTeX\ Class Files,~Vol.~14, No.~8, August~2015}%
{Shell \MakeLowercase{\textit{et al.}}: Bare Demo of IEEEtran.cls for IEEE Journals}
% The only time the second header will appear is for the odd numbered pages
% after the title page when using the twoside option.
%
% *** Note that you probably will NOT want to include the author's ***
% *** name in the headers of peer review papers.                   ***
% You can use \ifCLASSOPTIONpeerreview for conditional compilation here if
% you desire.

% If you want to put a publisher's ID mark on the page you can do it like
% this:
%\IEEEpubid{0000--0000/00\$00.00~\copyright~2015 IEEE}
% Remember, if you use this you must call \IEEEpubidadjcol in the second
% column for its text to clear the IEEEpubid mark.

% use for special paper notices
%\IEEEspecialpapernotice{(Invited Paper)}

% make the title area
%\IEEEpeerreviewmaketitle
\maketitle
%\IEEEpeerreviewmaketitle
% As a general rule, do not put math, special symbols or citations
% in the abstract or keywords.
\begin{abstract}
Describing the dimension reduction (DR) techniques by means of probabilistic models has recently been given special attention. Probabilistic models, in addition to a better interpretability of the DR methods, provide a framework for further extensions of such algorithms.  One of the new approaches to the probabilistic DR methods is to preserving the internal structure of data. It is meant that it is not necessary that the data first be converted from the matrix or tensor format to the vector format in the process of dimensionality reduction. In this paper, a latent variable model for matrix-variate data for canonical correlation analysis (CCA) is proposed. Since in general there is not any analytical maximum likelihood solution for this model, we present two approaches for learning the parameters. The proposed methods are evaluated using the synthetic data in terms of convergence and quality of mappings. Also, real data set is employed for assessing the proposed methods with several probabilistic and none-probabilistic CCA based approaches. The results confirm the superiority of the proposed methods with respect to the competing algorithms. Moreover, this model can be considered as a framework for further extensions.
\end{abstract}

% Note that keywords are not normally used for peerreview papers.

% For peer review papers, you can put extra information on the cover
% page as needed:
% \ifCLASSOPTIONpeerreview
% \begin{center} \bfseries EDICS Category: 3-BBND \end{center}
% \fi
%
% For peerreview papers, this IEEEtran command inserts a page break and
% creates the second title. It will be ignored for other modes.
%\IEEEpeerreviewmaketitle
%\onecolumn
%\twocolumn

\section{Introduction}

% The very first letter is a 2 line initial drop letter followed
% by the rest of the first word in caps.
%
% form to use if the first word consists of a single letter:
% \IEEEPARstart{A}{demo} file is ....
%
% form to use if you need the single drop letter followed by
% normal text (unknown if ever used by the IEEE):
% \IEEEPARstart{A}{}demo file is ....
%
% Some journals put the first two words in caps:
% \IEEEPARstart{T}{his demo} file is ....
%
% Here we have the typical use of a "T" for an initial drop letter
% and "HIS" in caps to complete the first word.
\IEEEPARstart{R}{ecently}, probabilistic interpretation of statistical dimension reduction techniques in the subspace domain has been applied in the different applications\cite{7206576,doi:10.1093/bioinformatics/btu134,SHARIFI2017638}.  Probabilistic dimension reduction models offer many benefits, including the handling of missing and outlier data\cite{CHEN20093706}, automatic selection of number of projection vectors\cite{bayesian-pca} and
the extending of the standard dimension reduction methods to more complex ones such as mixture models \cite{doi:10.1162/089976699300016728,6714575} or non-linear models \cite{Lawrence:2005:PNP:1046920.1194904,journals/jmlr/TitsiasL10}.\par
Tipping and Bishop  presented a probabilistic model for principal component analysis (PCA) called probabilistic PCA (PPCA) and showed that a relation could be found between the projections extracted by the PCA method and the maximum likelihood solution of an restricted factor analysis model \cite{tipping1999probabilistic}. Lawrence proposed a dual model for PPCA and extended it to the non-linear case through the gaussian processes \cite{Lawrence:2005:PNP:1046920.1194904}. Bach and Jordan presented probabilistic interpretation of canonical correlation analysis (CCA) and called it probabilistic CCA (PCCA)\cite{bach2005probabilistic}. In this method, a latent variable model is used to describe two gaussian random vectors. Recently different variations of PCCA have also been proposed \cite{Klami2013Bayesian,michaeli2016nonparametric,sarvestani2016ff}.\par

All the above mentioned methods assume that the input data or features are described as vectors. However, in many applications, we encounter with the data that have an intrinsic structure such as matrix or tensor. For example, in face recognition task, pixels of the image can be considered as the features, which have matrix structure, or in image processing, 2D Gabor functions are frequently used for feature extraction that the outputs of which have the matrix structure\cite{Daugman:85}. In traditional dimension reduction approaches, these structures are usually broken and the features are concatenated into a long vector. However, this leads to the small sample size problem and the increase in the computational cost due to the large matrices\cite{yang2004two}. Therefore, in recent years, attention has been paid to the use of algorithms that do not use the data transformation (converting data into a vector) as a preprocessing step. Two-dimensional PCA (2DPCA)\cite{yang2004two}, general low rank approximation of matrix (GLRAM)\cite{Ye2005} and two-dimensional CCA (2DCCA)\cite{lee2007two,Sun2010Two-dimensional} are among the first examples of these algorithms.\par The probabilistic interpretation of 2-or-more-dimensional subspace feature extraction techniques is also an active research field. Tao and et al. presented a decoupled bayesian tensor analysis model which could reduce the dimensions of tensor data and automatically determine the appropriate dimensions \cite{tao2008bayesian}. In \cite{Yu2011Matrix}, by applying matrix variate distributions \cite{b1}, a probabilistic higher-order PCA for matrix-variate data was introduced and variational expectation-maximization (EM) was applied for learning the parameters of the model. Another probabilistic model for 2DPCA was proposed by Zhao and et al. called bilinear PPCA (BPPCA) which extend the previous models by defining three separate model of noises called column, row and common noise. BPCCA formulated its proposed model with a two-stage representation and the parameters of the model were computed with both maximum likelihood estimation (MLE) and EM \cite{zhao2012bilinear}. Safayani and et al. introduced probabilistic 2DCCA (P2DCCA) in which two models on columns and rows of images called left and right probabilistic models were defined \cite{Safayani2017}. For learning the parameters, it is assumed that the parameters of right probabilistic model is known and the observations are projected to the corresponding latent spaces. Then the parameters of the left probabilistic model are estimated using EM algorithm. In a similar procedure, and parallel to the left probabilistic model, the parameters of right probabilistic model are learned. This procedure is repeated until convergence. In this approach, it is assumed that the columns of observation matrix are independent and its probability distribution is constructed by producting the probability of the corresponding columns. Also, because of existence of two models on the rows and the columns, a pair of data can not be mapped to a single latent space. This causes the model not to be a generative model.\par

In this paper, a probabilistic model for CCA with matrix data assumption is presented. In this model, two random matrices are related through a latent variable that has a matrix-variate normal distribution. Since there is no closed-form solution for learning the parameters of this model, two approaches are proposed. The first approach called unilateral matrix-variate CCA (UMVCCA) assumes that the latent variables are only projected from one side (rows or columns), and with this assumption, the model parameters are estimated using EM algorithm. In the second approach, the simple assumption of the first approach is not considered and the hidden variables are written from both sides (rows and columns). This model is named bilateral matrix-variate CCA (BMVCCA). To learn the parameters of this model, an algorithm based on the variational EM\cite{Jordan1999An} is proposed. In the learning algorithm, the posterior distribution is estimated using a matrix-variate normal distribution and a lower bound of the log-likelihood is maximized with respect to the variational parameters that are here mean matrix and two covariance matrices of the matrix-variate normal distribution. \par
The proposed algorithms are initially assessed using synthetic data for convergence analysis and also accuracy of mapping matrices, and then are evaluated on the "NIR-VIS 2.0" \cite{li2013casia} face database and their results are compared with competing algorithms such as CCA, 2DCCA\cite{lee2007two}, PCCA\cite{bach2005probabilistic} and P2DCCA\cite{Safayani2017}.
The rest of this paper is organized as follows: in Section \ref{secbackground}, matrix-variate normal distribution as well as CCA, 2DCCA and PCCA are briefly reviewed. The proposed method is presented in Section \ref{secMVCCA}. Section \ref{secexp} is devoted to the assessing of proposed methods using both syntectic and real data. Finally, the paper is concluded in Section \ref{secconc}.

\section{Related works}\label{secbackground}
\subsection{Matrix-variate normal distribution}
Since in two-dimensional probabilistic models we deal with random matrices (instead of random vectors) variables, it is convenient to use matrix-variate distributions to model them. In general, matrix-variate distributions are a 2D generalization of multivariate distributions \cite{b1}. Matrix-variate normal distribution $X \sim \mvgauss{M}{\Sigma}{\Phi} $ is the most famous one and is defined as follows:
\begin{equation} \label{eq0.1}
\frac{1}{(2\pi)^{\frac{1}{2}mn}\lvert\Sigma\rvert^{\frac{1}{2}n}\lvert\Phi\rvert^{\frac{1}{2}m}}exp\big[tr(-\frac{1}{2}\Sigma^{-1}(X-M)\Phi^{-1}(X-M)')\big],
\end{equation}
where $ X\in{\rm I\!R}^{m\times n} $ is a random matrix,  $ M\in{\rm I\!R}^{m\times n} $ is the mean matrix, $\Sigma\in{\rm I\!R}^{m\times m} \succ 0 $  and  $ \Phi\in{\rm I\!R}^{n\times n} \succ 0 $ are the column and row covariance matrices respectively and $tr(A)$ denotes trace of matrix $A$.
Also, it can be shown that if X follows $X \sim \mvgauss{M}{\Sigma}{\Phi} $ then $ vec(X) \sim \gauss{vec(M)}{\Sigma \otimes \Phi}$, where $\otimes$ is the kronecker product \cite{b1}.
\subsection{Canonical Correlation Analysis (CCA)}
Canonical correlation analysis is a method that seeks to find relationships between two multivariate sets of variables\cite{hotelling1936relations}. CCA maps each set of data to a common subspace in which the correlation between the two data sets is maximized. For example assumes that  $x^1\in {\rm I\!R}^{m^1}$ and  $x^2\in {\rm I\!R}^{m^2}$ are two random vectors. CCA finds two linear mappings ${w^1}' x^1$ and ${w^2}' x^2$ in which the following criteria is maximized
\begin{align}
\label{2dcca_opt}
\argmax_{w^1,w^2}& \enskip cov\Big({w^1}'x^1,{w^2}'x^2\Big), \\ \nonumber
&s.t. \quad var({w^1}'x^1)=1,    \\ \nonumber
&\enskip \enskip \enskip \enskip \enskip var({w^2}'x^2)=1.
\end{align}
% \begin{equation} \label{p1}
%\rho = corr(({w^1})^Tx^1,({w^2})^Tx^2)=\frac{({w^1})^TC_{12}w^2}{\sqrt{({w^1})^TC_{11} w^1}\sqrt{({w^2})^TC_{22} w^2}}.
%\end{equation}
 It can be shown that the optimal $w^1$ and $w^2$ can be obtained by solving the following eigen problems:
\begin{equation} \label{cca-1}
{C_{11}}^{-1}C_{12}{C_{22}}^{-1}C_{21}w^1=\lambda^2 w^1,
\end{equation}
\begin{equation} \label{cca-2}
{C_{22}}^{-1}C_{21}{C_{11}}^{-1}C_{12}w^2=\lambda^2 w^2,
\end{equation}
where $C_{11}$ and $C_{22}$ are the autocovariance matrices of random vectors $x^1$ and $x^2$ respectively and $C_{12}$ is the cross-covariance matrix of them and $\lambda^2$ is the largest eigenvalue and is equal to the square of canonical correlations.
\subsection{Two-dimensional CCA (2DCCA) }
2DCCA is a 2D extension of CCA which works directly on matrix data\cite{lee2007two} . Let $\{X_n^j\in {\rm I\!R}^{m^j \times n^j}|_{j=1}^2, n=1,...,N\}$ are realizations of random matrix variable $X^j|_{j=1}^2$. Without loss of generality, here, we assume that the random variables are zero mean. 2DCCA tries to obtain projection vectors $l^j|_{j=1}^2$ and  $r^j|_{j=1}^2$ that maximizes the following optimization problem:
\begin{align}
\label{2dcca_opt}
\argmax_{l^1,r^1,l^2,r^2}& \enskip cov\Big({l^1}'X^1r^1,{l^2}'X^2r^2\Big), \\ \nonumber
&s.t. \quad var({l^1}'X^1r^1)=1,    \\ \nonumber
&\enskip \enskip \enskip \enskip \enskip var({l^2}'X^2r^2)=1.
\end{align}
Since there is no closed-form solution for this problem, 2DCCA assumes that once $r^j|_{j=1}^2$ are fixed and after some simplifications the optimization formula converts to
\begin{align}
\label{2dcca_opt1}
&\argmax_{l^1,l^2} {l^1}'\Sigma_{12}^r{l^2}, \\ \nonumber
 &s.t.  \enskip {l^1}'\Sigma_{11}^rl^1=1,   \\ \nonumber
 & \quad\quad {l^2}'\Sigma_{22}^rl^2=1,
\end{align}
where $\Sigma^r_{ij}=\frac{1}{N}\sum_{n=1}^NX^i_nr^1{r^2}'{X^j_n}'$, and once again the $l^j|_{j=1}^2$ are assumed to be fixed and the following formula is obtained:
%\begin{align} \label{bam2.1}
%\Sigma^r_{ij}&=\frac{1}{N}\sum_{n=1}^NX^i_nr^1{r^2}'{X^j_n}',
%\end{align}
 \begin{align}
\label{2dcca_opt2}
&\argmax_{r^1,r^2} {r^1}'\Sigma_{12}^l{r^2}, \\ \nonumber
&s.t.  \enskip {r^1}'\Sigma_{11}^lr^1=1,   \\ \nonumber
 &\quad\quad {r^2}'\Sigma_{22}^lr^2=1,
\end{align}
where $\Sigma^l_{ij}=\frac{1}{N}\sum_{n=1}^N{X^i_n}'l^1{l^2}'X^j_n$. By solving (\ref{2dcca_opt1}) and (\ref{2dcca_opt2}) iteratively until convergence, the optimum transformations $\{l^1,r^1,l^2,r^2\}$ are obtained. It can be shown that optimization (\ref{2dcca_opt1}) and (\ref{2dcca_opt2}) converts to the following eigen problems:
\begingroup
\begin{align}
\label{bam-2}
&\begin{bmatrix}
0 & \Sigma^r_{12} \\
\Sigma^r_{21} & 0 \\
\end{bmatrix}
\begin{bmatrix}
l^1 \\
l^2 \\
\end{bmatrix}
=\lambda
\begin{bmatrix}
\Sigma^r_{11} &     0        \\
0             & \Sigma^r_{22} \\
\end{bmatrix}
\begin{bmatrix}
l^1 \\
l^2 \\
\end{bmatrix}, \\ %%%%%%
\label{bam-1}
&\begin{bmatrix}
0 & \Sigma^l_{12} \\
\Sigma^l_{21} & 0 \\
\end{bmatrix}
\begin{bmatrix}
r^1 \\
r^2 \\
\end{bmatrix}
=\lambda
\begin{bmatrix}
\Sigma^l_{11} &     0        \\
0             & \Sigma^l_{22} \\
\end{bmatrix}
\begin{bmatrix}
r^1 \\
r^2 \\
\end{bmatrix}.
\end{align}
\endgroup
The $d_1$ largest eigenvectors of (\ref{bam-2}) generates the columns of matrix $L^j|_{j=1}^2$, similarly $d_2$ largest eigenvectors of (\ref{bam-1}) determines columns of matrix $R^j|_{j=1}^2$.

\subsection{Probabilistic CCA (PCCA)}
PCCA defines the following latent variable model\cite{bach2005probabilistic}:
\begin{equation}\label{EQ4}
x^{j}=W^{j}z+\mu^{j}+\epsilon^{j}, \qquad j\in \{1,2\},
\end{equation}
where $x^j$ is the observation random vector, $z$ is the latent vector with a multivariate normal distribution with zero mean and identity covariance matrix, $\epsilon_{j}$ is the residual noise vector which follows multivariate normal distribution with expectation of zero and $\Psi^j$ covariance matrix and $\mu^{j}$ is the mean vector of random vector $x^j$. In this model, conditioned on the latent variable $z$, $x^1$ and $x^2$ are independent. It can be shown that the following distributions result from (\ref{EQ4}):
\begin{align} \label{eq0.385}
p(x‍‍^j|z) &= \gauss{W^jz+\mu^j}{\Psi^j}, \\ \label{eq0.5}
p(x|z) &= \gauss{Wz+\mu}{\Psi}, \\ \label{eq}
p(x) &= \gauss{\mu}{
\Sigma
},
\end{align}
where $x=[{x^1}',{x^2}']'$, $W=[{W^1}',{W^2}']'$, $\mu=[{\mu^1}',{\mu^2}']'$, $\Sigma=\begin{bmatrix}
W^1{W^1}'+\Psi^1 & W^1{W^2}' \\
W^2{W^1}'        & W^2{W^2}'+\Psi^2
\end{bmatrix}$ and $\Psi=[{\Psi^1}',{\Psi^2}']'$. Let $x^1_n|_{n=1}^N$ and  $x^2_n|_{n=1}^N$ denote a set of observation vectors. The maximum log likelihood estimates of parameters $\theta=(W,\Psi)$ can be obtained by maximizing
\begin{align} \label{eq0.385}
\mathcal{L}(\theta)=\frac{N}{2}\log\Sigma+\frac{1}{2}\sum_{n=1}^Ntr\Sigma^{-1}(x_n-\mu){(x_n-\mu)}'+const,
\end{align}
which leads to
\begin{alignat}{2}\label{bam14}
W_{ML}^j &= \widetilde{\Sigma}_{j j}U^j_{d}S^j,&&\quad j\in \{1,2\},\\ \label{bam15}
\Psi_{ML}^j &=\widetilde{\Sigma}_{j j}-W_{ML}^j{W_{ML}^j}', && \quad j\in \{1,2\},
\end{alignat}
where $S^1$ and $S^2$ are the arbitrary matrices such that $S^1{S^2}'=C_d$ and $C_d$ is the diagonal matrix of the first $d$ canonical directions, and $U^j_d$ consists of first $d$ canonical directions and $\widetilde{\Sigma}_{j j}$ is the sample covariance matrix of $x^j$.\\
In \cite{bach2005probabilistic}, also an iterative algorithm based on EM was proposed to maximize (\ref{eq0.385}). For this reason, $z_n|_{n=1}^N$ are considered as the missing values in the optimization algorithm and the complete log-likelihood is as follows:
\begin{equation} \label{bam15}
\mathcal{L}(\theta) = \sum_{n=1}^{N} \{\ln P(x_n|z_n)+\ln P(z_n)\}.
\end{equation}
By substituting (\ref{eq0.5}) into  (\ref{bam15}) and some mathematics, the expected value of log-likelihood function with respect to the posterior distribution is obtained :
\begin{align}\label{eq8}
%E[\ln P(X_1,X_2,Z)] = \sum_{i=1}^{N} \{E[\ln \gauss{Wz_i+\mu}{\Psi}]+E[\ln \gauss{0}{I_d}]\}
Q(\theta|\theta^{(t)})&=\sum_{n=1}^{N} \Big\lbrace-\frac{1}{2}|\Psi |-\frac{1}{2}(x_n'\Psi^{-1}x_n)-\frac{1}{2}(\langle z_nz_n'\rangle)\\\nonumber
&-\frac{1}{2}(W'\Psi^{-1}W\langle z_nz_n'\rangle)+\langle z_n \rangle W'\Psi^{-1}x_n \Big\rbrace,
\end{align}
where
\begin{align} \label{eq2.9}
\langle z_n \rangle&=MW'\Psi^{-1}x_n, \\ \label{eq2.10}
\langle z_nz_n'\rangle&= M + \langle z_n \rangle\langle z_n \rangle',\\\label{eq2.11}
M&=(W'\Psi W+I)^{-1}.
\end{align}
By maximizing (\ref{eq8}) with respect to the parameters and substituting (\ref{eq2.9}) and (\ref{eq2.10}) into the solutions, the final update formulas are obtained as follows:
\begingroup
\renewcommand*{\arraystretch}{2.2}
\begin{align}
\label{bam16}
W_{t+1}=&\widetilde{\Sigma}\Psi_t^{-1}W_tM_t(M_t+M_tW_t'\Psi_t^{-1}\widetilde{\Sigma}\Psi_t^{-1}W_tM_t)^{-1}, \\
%\Psi_{t+1} =&
%\begin{bmatrix}
%(\widetilde{\Sigma}-\widetilde{\Sigma}\Psi_t^{-1}W_tM_tW_{t+1})_{11} &                               0                                       \\
%\label{bam17}
%                              0                                      & (\widetilde{\Sigma}-\widetilde{\Sigma}\Psi_t^{-1}W_tM_tW_{t+1})_{22}
%\end{bmatrix}.
\Psi_{t+1} =&\widetilde{\Sigma}-\widetilde{\Sigma}\Psi_t^{-1}W_tM_tW_{t+1}.
%\begin{bmatrix}
%(\widetilde{\Sigma}-\widetilde{\Sigma}\Psi_t^{-1}W_tM_tW_{t+1})_{11} &                               0                                       \\
%\label{bam17}
%                              0                                      & (\widetilde{\Sigma}-\widetilde{\Sigma}\Psi_t^{-1}W_tM_tW_{t+1})_{22}
%\end{bmatrix}.
\end{align}
\endgroup

\section{Latent variable model for two-dimensional CCA}\label{secMVCCA}

Inspired by \cite{bach2005probabilistic,Yu2011Matrix}, the proposed model extends PCCA for matrix data as follows:
\begin{equation} \label{eq1}
X^{j} = L^jZ{R^j}' + M^j + \Xi^{j},\qquad j\in \{1,2\},
\end{equation}
where $j$ is the index of observation variable, $X^j\in {\rm I\!R}^{m^j \times n^j}$ and $\Xi^j \in {\rm I\!R}^{m^j \times n^j}$ are observed variable and residual noise matrix respectively, $Z\in {\rm I\!R}^{d_1 \times d_2}$ is the latent matrix variable, $M^j\in {\rm I\!R}^{m^j \times n^j}$ indicates means of corresponding observed variable ( without loss of generality from here we assume that the observed variables are zero mean) and $L‍‍‍‍^j\in {\rm I\!R}^{m^j\times d_1}$ and  $R^j\in {\rm I\!R}^{n^j\times d_2}$  are the left and right projection matrices respectively.  The latent and noise matrix variables have the following distributions:
\begin{align}
p(Z)& = \mvgauss{0}{I}{I}, \label{eq1.0.1} \\
 \label{bam39}
p(\Xi^j)& = \mvgauss{0}{\Psi^j_L}{\Psi^j_R},  \quad \Psi^j_L , \Psi^j_R \succeq 0, j\in \{1,2\},
\end{align}
where $ \Psi^j_L \in {\rm I\!R}^{m^j \times m^j}$ and $ \Psi^j_R \in {\rm I\!R}^{n^j \times n^j}$ are the positive-semidefinite covariance matrices of noise matrix variables.

From the linear model presented in equation (\ref{eq1}) and the noise distributions in equation (\ref{bam39}), the conditional distribution of observation variables given the latent matrix are obtained as:
\begin{equation} \label{eq1.1}
p(X^{j}|Z)= \mvgauss{L^{j}Z{R^{j}}'}{\Psi^{j}_{L}}{\Psi^{j}_{R}},\quad j\in \{1,2\}.%\\
%X^{2}_{i}|Z_{i}\sim \mvgauss{L^{2}Z_{i}R^{2}}{\Psi^{2}_{L}}{\Psi^{2}_{R}}
\end{equation}
%For estimating the parameters the Likelihood of the model is obtained as follows:
%\begin{equation} \label{t1}
%L = \prod_{i=1}^{N} P(X^{1}_{i},X^{2}_{i},Z_{i})
%\end{equation}
%Where, $N$ indicates the number of training data.
This model is an extension to the PCCA \cite{bach2005probabilistic}, the difference is that here the observed, latent and noise variables are represented as matrices instead of vectors. It can be shown that the vector form of (\ref{eq1}) is obtained as follows
\begin{alignat}{2} \label{eq1_c}
vec(X^{j}) = W^jvec(Z)& + vec(M^j) \\\nonumber
 &+ vec(\Xi^{j}), &\enskip j\in \{1,2\},\\
p(vec(Z))& = \mathcal{N}({0},{I}) \label{eq1.0.10}, &\\ \label{bam391}
p(vec(\Xi^j))& = \mathcal{N}({0},{\Psi^j_L}),  \quad  \Psi^j \succeq 0,  &\quad j\in \{1,2\},
\end{alignat}
%where
%\begin{align}
%p(vec(Z))& = \mathcal{N}({0},{I}) \label{eq1.0.1} \\ \label{bam391}
%p(vec(\Xi^j))& = \mathcal{N}({0},{\Psi^j_L})  \quad  \Psi^j \succeq 0  \qquad  for \quad j=1,2
%\end{align}
where $vec(.)$ is an operator that vectorize the input matrix by concatenating its columns, $W^j=(R^j\otimes{L^j})$ and $\Psi^j=\Psi^j_R\otimes \Psi^j_L$. As it can be observed projection matrix $W_j$ is the  Kronecker product of right and left projection matrices and similarly noise covariance matrix $\Psi^j$ can be decompose into right and left noise covariance matrices. Therefore, the number of free parameters of model (\ref{eq1_c}) is much less than those of model (\ref{EQ4}).\par

The joint probabilistic distribution for a data pair and the corresponding subspace respresentation could be written as
\begin{equation} \label{t0}
P(X^{1},X^{2},Z) = P(X^{1}|Z)P(X^{2}|Z)P(Z).
\end{equation}
While in PCCA the likelihood function of observed data, i.e., $P(X^1,X^2)$ could be obtained via integrating out the latent variable, here due to the use of matrix-variate distributions, in the general case $P(X^1,X^2)$ and also posteriori distribution $P(Z|X^1,X^2)$ do not follow a matrix-variate normal distribution. Therefore, we propose two approaches for learning the parameters. In the first one, we simplify the model by assuming only one projection matrix (left or right) so that the posteriori distribution can be derived based on a matrix-variate normal distribution and a solution based on EM algorithm can be provided. We call this approach unilateral matrix variate CCA model(UMVCCA). In another approach that we call it bilateral matrix variate CCA model (BMVCCA), we consider the model in general case (with both left and right projections) and estimate the posterior distribution $ P(Z|X^1,X^2)$ using a parametric matrix-variate normal distribution $q(Z)$ and a lower bound of the log-likelihood is maximized using variational EM algorithm \cite{Jordan1999An}. Each of these approaches is discussed here.

\subsection{Unilateral matrix variate CCA (UMVCCA)}
We assume that one of the left or right mapping matrices(here, without loss of generality the left mapping matrix) in (\ref{eq1}) are replaced by the identity matrix therefore we have
\begin{align} \label{bamp1}
X‍‍‍^j &= {ZR^j}' + \Xi^j,\quad j\in \{1,2\},\\
p(\Xi^j) &= \mvgauss{0}{I}{\Psi_{R}^{j} },
\end{align}
%\begin{equation} \label{bamp2}
%  p(\Xi^j) = \mvgauss{0}{I}{\Psi_{R}^{j} }
%\end{equation}
where $Z\in {\rm I\!R}^{m^j \times d_2}$. With assumption of $m^1=m^2=m$, the above models can be reformulated into one factor analysis model as follows:
\begin{equation} \label{bamp1}
X‍‍‍ = {ZR}' + \Xi,   \\
\end{equation}
where $X=[X^1,X^2]\in{\rm I\!R}^{m \times (n^1+n^2)}$, $R=[{R^1}',{R^2}']'\in{\rm I\!R}^{(n^1+n^2) \times d_2}$ and $\Xi=[\Xi^{1},\Xi^{2}]\in{\rm I\!R}^{m \times (n^1+n^2)}$. Following distributions can be obtained:
\begin{equation} \label{bamp2}
  p(\epsilon) = \mvgauss{0}{I}{\big[ \begin{smallmatrix} \Psi_{R}^{1} & 0 \\ 0 & \Psi_{R}^{2}\end{smallmatrix} \big]},
\end{equation}
\begin{equation} \label{pdf1}
p(X|Z) = \mvgauss{ZR'}{I}{\Psi_{R}},
\end{equation}
\begin{equation} \label{pos1}
P(Z|X) = \mvgauss{SR'\Psi_{R}^{-1}X'}{I}{S},
\end{equation}
\begin{equation} \label{Update_S}
S=(R'\Psi_{R}^{-1}R+I)^{-1},
\end{equation}
where $ \Psi_{r}=\big[ \begin{smallmatrix} \Psi_{R}^{1} & 0 \\ 0 & \Psi_{R}^{2}  \end{smallmatrix} \big] $.% and $S=(R'\Psi_{R}^{-1}R+I)^{-1}$.

For learning the parameters of UMVCCA, we use well-known EM algorithm. Let $X_n^1|_{n=1}^{N}$ and $X_n^2|_{n=1}^{N}$ be defined as $N$ pairs of training data. $\{X_n,Z_n\}_{n=1}^{N}$ is the complete data and the complete log likelihood is:
\begin{equation} \label{ulikelihood}
\mathcal{L}(\theta) = \Sigma_{n=1}^{N}ln\{P(X_n,Z_n)\}  =\Sigma_{n=1}^{N}ln\{P(X_n|Z_n)P(Z_n)\}.   \\
\end{equation}
%In this case the posteriori distribution can be derived as
%\begin{equation} \label{pos_u}
%P(Z_n|X_n) = \mvgauss{SR'\Psi_{R}^{-1}X_n'}{I}{S}
%\end{equation}
%\begin{equation} \label{pos_u_S}
%S=(R'\Psi_{R}^{-1}R+I)^{-1}
%\end{equation}
The rest of the derivations is straight forward. The final EM update formulas are
%\begin{align} \label{apt6}
%   \langle Z_i\rangle    &=  X_{i}\Psi_{R}^{-1}RS  \\ \label{apt7}
%  \langle Z_i'Z_i\rangle &= S + \langle Z_i'\rangle \langle Z_i\rangle
%\end{align}
\begin{align} \label{uupdate_R}
R^{*} &=\tilde{\Sigma}\Psi^{-1}_{R}RS[mS+SR'\Psi^{-1}_{R}\tilde{\Sigma}\Psi^{-1}_{R}RS]^{-1},\\ \label{uupdate_si}
\Psi_{R}^{*} &=\frac{1}{m}\tilde{\Sigma}-\frac{2}{m}RSR'\Psi^{-1}_{R}\tilde{\Sigma}+\frac{1}{m}RSR'\Psi^{-1}_{R}\tilde{\Sigma}\Psi^{-1}_{R}RSR',
\end{align}
where $\tilde{\Sigma} =\frac{1}{N} \sum_{i=1}^N X_i'X_i  $ is data scatter matrix and $N$ is the number of training data. Algorithm (\ref{algo1}) shows the steps of this algorithm.
 \begin{algorithm}
\renewcommand{\algorithmicrequire}{\textbf{Input:}}
\renewcommand{\algorithmicensure}{\textbf{Output:}}
\caption{\text{UMVCCA algorithm}}
 \begin{algorithmic}[1]
 \REQUIRE $X_n^1|_{n=1}^{N}$ and $X_n^2|_{n=1}^{N}$, initialization of $R^j$ with random matrices and $\Psi_{R}^{j}$ with identity matrices, for\quad j=1,2
  \REPEAT
  \STATE
   Update $R$ and $\Psi_R$  using (\ref{uupdate_R}) and (\ref{uupdate_si}).
  \UNTIL{ change of $\mathcal{L}$ is smaller than a threshold}
  \ENSURE  $R^j$ and $\Psi_{R}^{j}$, for\quad j=1,2
 \end{algorithmic}
 \label{algo1}
 \end{algorithm}
%which are very similar to EM update formulas in PCCA

\subsection{Bilateral matrix variate CCA (BMVCCA)}
In this case, we assume that in equation (\ref{eq1}) both left and right projection matrices are exist. For maximizing the likelihood function, we need to estimate the posterior of latent variable $Z_n$ given the observed variables $X^1_n$ and $X^2_n$. However, in general there is no matrix-variate formulation for this posterior. Therefore we employ variational-EM algorithm \cite{Jordan1999An} for maximizing the lower-bound of the likelihood function. In variational-EM, a parameterised variational distribution $q(Z_n)$ is chosen to estimate the posteriori distribution $P(Z|X^1,X^2)$ and then its parameters are optimised. Here, we consider the following parametric distribution in the form of matrix-variate normal for $q(Z_n)$:
\begin{equation} \label{var4.0.2}
q(Z_{n})\sim \mvgauss{C_{n}}{O}{S},
\end{equation}
where $ C_n\in {\rm I\!R}^{d_1 \times d_2}$ is the mean matrix and $ O\in {\rm I\!R}^{d_1 \times d_1}$ and $ S\in {\rm I\!R}^{d_2 \times d_2}$ are the column and row covariance matrices respectively. The variational parameters are optimised to maximized a lower bound $ \mathcal{L}(q) $  of the data log-likelihood function which can be written as
\begin{align} \label{var4}
 \mathcal{L}(q)=\sum_n \int \ln\biggl[ \frac{P(X_{n}^{1},X_{n}^{2},Z_{n})}{q(Z_{n})} \biggl] q(Z_{n}) dZ_{n} = \\\nonumber
 \sum_n \mathbb{E}_q\Biggl[\ln\biggl[ \frac{P(X_{n}^{1},X_{n}^{2},Z_{n})}{q(Z_{n})}\biggl] \Biggl],
\end{align}

%\begin{equation} \label{app2:1}
%\mathcal{L}(q) =
% %\sum_i \mathbb{E}_{q}\Big[\ln\big[ \frac{P(X_{i}^{1},X_{i}^{2},Z_{i})}{q(Z_{i})}\big] \Big] =
% \sum_n \mathbb{E}_{q}\Big[\ln\big[ \frac{P(X^{1}_{n},X^{2}_{n}|Z_{n})P(Z_{n})}{q(Z_{n})}\big] \Big],
%\end{equation}
where $ \mathbb{E}_{q} $ stands for "expected w.r.t the variational distribution". The optimization of the variational parameters $\{C_n,O,S\}$ yields the following update formulas (variational E-step).
\begin{equation}\label{var4.1}
O^{*} = \biggl[ \frac{1}{d_{2}}\sum_{j=1}^{2}tr[ {R^{j}}'{\Psi^{j}_{R}}^{-1}R^{j}S ]{L^{j}}'{\Psi^{j}_{L}}^{-1}L^{j} + \frac{1}{d_{2}}tr[S]\times I \biggl]^{-1},
\end{equation}
\begin{equation}\label{var4.2}
S^{*} = \biggl[ \frac{1}{d_{1}}\sum_{j=1}^{2}tr[ {L^{j}}'{\Psi^{j}_{L}}^{-1}L^{j}O ]{R^{j}}'{\Psi^{j}_{R}}^{-1}R^{j} + \frac{1}{d_{1}}tr[O]\times I \biggl]^{-1},
\end{equation}
\begin{align}
  \label{var5}
vec(C_{n}^*) = \Biggl[ \sum_{j=1}^{2}\biggl[{R^{j}}'{\Psi^{j}_{R}}^{-1}R^{j}\otimes {L^{j}}'{\Psi^{j}_{L}}^{-1}L^{j}\biggl] + I\Biggl]^{-1}\times\\ \nonumber
vec\Biggl( \sum_{j=1}^{2}{L^{j}}'{\Psi^{j}_{L}}^{-1}X_{n}^{j}{\Psi^{j}_{R}}^{-1}R^{j} \Biggl).
\end{align}

%where $tr(A)$ means trace of matrix A. The optimization of mean of the variational distribution is not straight forward and like in \cite{yu2011matrix} it is optimized by solving a big linear equation
%\begin{align}
%  \label{var5}
%\Biggl[ \sum_{j=1}^{2}\biggl[{R^{j}}'{\Psi^{j}_{R}}^{-1}R^{j}\otimes {L^{j}}'{\Psi^{j}_{L}}^{-1}L^{j}\biggl] + I\Biggl]vec(C_{i}^*) = \\ \nonumber
%vec\Biggl( \sum_{j=1}^{2}{L^{j}}'{\Psi^{j}_{L}}^{-1}X_{i}^{j}{\Psi^{j}_{R}}^{-1}R^{j} \Biggl)
%\end{align}
After estimating the variational distribution $q(Z_n)$ in the E-step process, the update formulas for the rest parameters of the model are obtained as follows (variational M-step):
\begin{align} \label{var5.0.5}
{\Psi^j_L}^* = \biggl[ \frac{1}{Nn_j}P_L^j + \frac{1}{n_j}tr[{R^j}'{\Psi^j_R}^{-1}R^jS](L^jO{L^j}')\biggl],
\end{align}

\begin{align} \label{var5.0.6}
{\Psi^j_R}^* = \biggl[ \frac{1}{Nm_j}P + \frac{1}{m_j}tr[{L^j}'{\Psi^j_L}^{-1}L^jO](R^jS{R^j}')\biggl],
\end{align}
\begin{align} \label{var5.0.75}
{L^j}^* = \biggl[-\sum_{n=1}^NX_n^j{\Psi^j_R}^{-1}{R^j}'C_n'\biggl]
\biggl[ -Ntr[R^j{\Psi_R^j}^{-1}{R^j}'S]O \\ \nonumber
-\sum_{n=1}^N C_nR^j{\Psi_R^j}^{-1}{R^j}'{C_n}'\biggl]^{-1},
\end{align}
\begin{align} \label{var5.1}
{R^j}^* = \biggl[-\sum_{n=1}^N{C_n}'{L^j}'{\Psi^j_L}^{-1}X^j_n\biggl]
\biggl[ -Ntr[{L^j}'{\Psi_L^j}^{-1}L^jO]S \\ \nonumber
-\sum_{n=1}^N C_n'{L^j}'{\Psi_L^j}^{-1}L^jC_n\biggl]^{-1},
\end{align}
The algorithm of BMVCCA is presented in algorithm (\ref{algo2}).
 \begin{algorithm}
\renewcommand{\algorithmicrequire}{\textbf{Input:}}
\renewcommand{\algorithmicensure}{\textbf{Output:}}
\caption{\text{BMVCCA algorithm}}
 \begin{algorithmic}[1]
 \REQUIRE $X_n^1|_{n=1}^{N}$ and $X_n^2|_{n=1}^{N}$, initialization of $L^j$ and $R^j$ with 2DCCA algorithm and $\Psi_{L}^{j}$ $\Psi_{R}^{j}$ with identity matrices, for\quad j=1,2
  \REPEAT
  \STATE
  \textbf{E-STEP:}
  \STATE
   Update $O$, $S$ and $C_n$ using (\ref{var4.1}), (\ref{var4.2}) and (\ref{var5}) respectively.
   \STATE
   \textbf{M-STEP:}
   \STATE
   Update $\Psi_L^j$, $\Psi_R^j$ $L^j$ and $R^j$ using (\ref{var5.0.5}), (\ref{var5.0.6}), (\ref{var5.0.75}) and (\ref{var5.1}) respectively for\quad j=1,2.
   \UNTIL{ change of $\mathcal{L}$ is smaller than a threshold}
  \ENSURE  $L^j$,$R^j$,$\Psi_{L}^{j}$ and $\Psi_{R}^{j}$ for\quad j=1,2
 \end{algorithmic}
 \label{algo2}
 \end{algorithm}
%which are very similar to EM update formulas in PCCA

\subsection{Dimension Reduction}
Observation data matrices can be projected into the low-dimensional space using $\{L^j,R^j\}_{j=1}^2$. But it is more natural to use probabilistic projection. Here, similar to PPCA, we represent the observed data into the low-dimensional space by using the mean of posterior distribution,i.e., $E(Z|T^1,T^2)$. However, in BMVCCA, $q(Z)$ estimates the posterior distribution so we consider the mean of $q(Z)$. As it can be seen this equation can be applied whenever we have two observation matrices. In the case of one observation matrix (here, $X^1$), we consider $E[q(Z|X^1,M^2)]$  as the low-dimensional representation of $X^1$, where $M^2$ is the average of $X^2$ random matrix in the training dataset. Similar formula can be obtained for $X^2$.

\section{Experiments}\label{secexp}
In this section we analyze the performance of algorithm using both synthetic and real data.
\subsection{ Synthetic data}
\subsubsection{Convergence of Algorithm}\label{syn_conv}
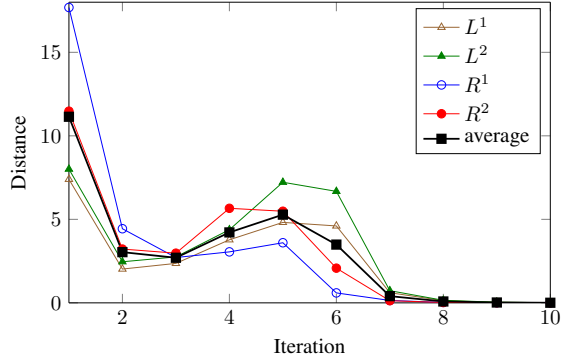
\begin{figure}[t!]
\begin{center}
%\begin{latin}
\input{"fig_synthetic_data_convergence"}
%\end{latin}
\caption{Distance between two successive iteration mappings of BMVCCA}
\label{fig4:1ffff}
\end{center}
\end{figure}

In this section, the convergence of the algorithm is analyzed using synthetic data. For generating the synthetic data, we at first sample each element of the projection matrices ${L^j}\in {\rm I\!R}^{32\times 15}|_{j=1}^2$ and ${R^j}\in {\rm I\!R}^{32\times 15}|_{j=1}^2$ from the uniform distribution in the range of zero and one. Then for each pair of observed data, the latent matrix $Z \in {\rm I\!R}^{15\times 15}$ and residual matrices $\Xi^{j}\in {\rm I\!R}^{32\times 32}|_{j=1}^2$ are sampled from $\mvgauss{0}{I}{I}$ and $\mvgauss{0}{0.1I}{0.1I}$ respectively and the observed data $\{ X^j \in {\rm I\!R}^{32\times 32}\}_{j=1}^{2}$ are generated using equation (\ref{eq1}). We generate 1000 samples with this procedure and run BMVCCA algorithm and compute the Frobenius norm of each projection matrix in different iterations of the algorithm. Fig. (\ref{fig4:1ffff}) plots the difference of computed norms in the successive iterations. As it can be seen the algorithm converges after some iterations.\\
\subsubsection{Analyzing the probabilistic subspace}
In this section, we want to get the mean of distribution of latent variables conditioned on the observed matrices and compute their distance to the true latent variables. Similar to the previous section, we generate 1000 pairs of data $\{ X^j \in {\rm I\!R}^{32\times 32}\}_{j=1}^{2}$ using equation (\ref{eq1}). This time, the projections ${L^j}\in {\rm I\!R}^{32\times 1}|_{j=1}^2$ and ${R^j}\in {\rm I\!R}^{32\times 1}|_{j=1}^2$ are vectors and come from the uniform distribution in the range of zero and one, for each data $Z \in {\rm I\!R}^{1\times 1}$ is a scalar sampled from $\mathcal{N}(0,1)$ distribution and residual matrices $\Xi^{j}\in {\rm I\!R}^{32\times 32}|_{j=1}^2$ are sampled from $\mvgauss{0}{0.1I}{0.1I}$.\\

We run BMVCCA to obtain $C_n|_{n=1}^{1000}$, which are the estimated mean of posterior distribution of $P(Z_n|X_n^1,X_n^2)$, in each iteration of algorithm. Then we compute the Euclidean distance between the true latent space $Z_n|_{n=1}^{1000}$ with the corresponding $C_n|_{n=1}^{1000}$ and plot it in Fig. (\ref{fig4:2}) for different iterations. As it can be observed the error is reduced and goes to zero. To investigate the effect of the number of training data, we repeat this experiment with different training samples ranging from 10 to 1000 samples. Fig. (\ref{fig4:1}) illustrates the result as it can be observed the accuracy of estimation is improved with increased training size.\\
 It should be noted that this experiment is applicable only for $1\times 1$ latent space. It is due to the fact that the obtained latent space leads to the true latent space up to a rotation and in general there is no solution for obtaining the rotation matrices. But it can be shown that whenever the subspace is restricted to $1\times 1$, there is no rotational matrix and the true and learned latent variables are equal up to a scaling factor which its magnitude can be removed by normalizing both latent spaces. However, its sign cannot be removed by this procedure and for canceling the sign of scaling factor, we compute Euclidean distance between the true latent space with both positive and negative sign of learned subspace and the minimum is chosen.

\begin{figure}[t!]
\centering
\begin{center}
\input{"fig_dist2"}
\caption{Distance between estimate and true latent space in varying iterations for BMVCCA}
\label{fig4:2}
\end{center}
\end{figure}
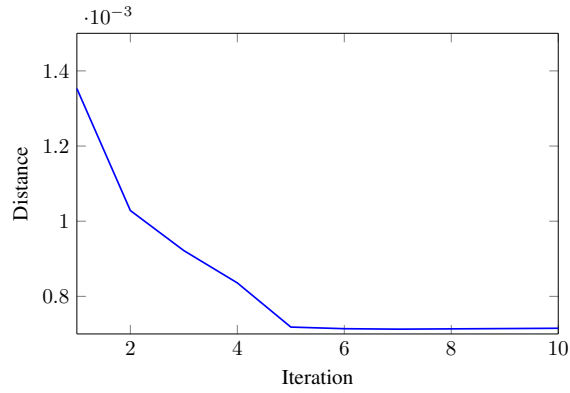

\begin{figure}[t!]
\begin{center}
\input{"fig_dist_to_true_post"}
\caption{Distance between estimate and true latent space with varying training samples for BMVCCA}
\label{fig4:1}
\end{center}
\end{figure}
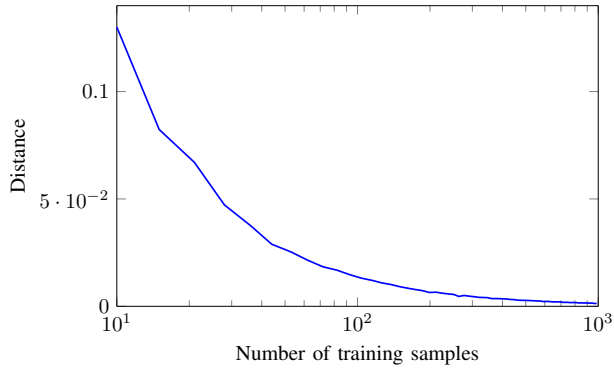

%But this time  we assume that for each data $Z \in {\rm I\!R}^{1\times 1}$ is a scalar sampled from $\mathcal{N}(0,1)$ distribution, each element of projection vectors ${L^j}\in {\rm I\!R}^{32\times 1}|_{j=1}^2$ and ${R^j}\in {\rm I\!R}^{32\times 1}|_{j=1}^2$ comes from the uniform distribution in the range of zero and one, and residual matrices $\Xi^{j}\in {\rm I\!R}^{32\times 32}|_{j=1}^2$ are sampled from $\mvgauss{0}{0.1I}{0.1I}$.
\subsubsection{Analyzing the learned projections of UMVCCA}
\begin{figure}
\centering

\input{myfile}
\caption{Illustration of the true and learned projection vectros of UMVCCA. Horizontal axis is the dimension of the vector ranging from 1 to 32. Left: true projection vectors, Right: learned projection vectors}
\label{mode1}
\end{figure}
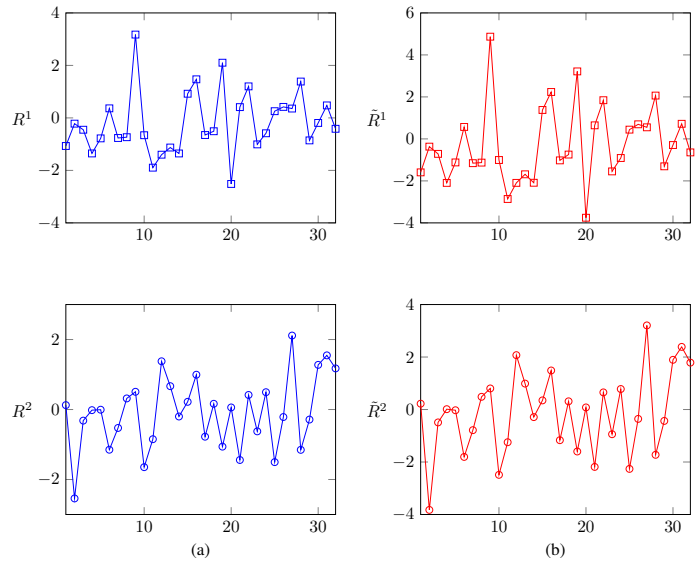

In this section, we try to examine the learned projection matrices of UMVCCA by using the synthetic data.  We generate 1000 pairs of observation data $\{ X^j \in {\rm I\!R}^{32\times 32}\}_{j=1}^{2}$ with the procedure similar to the previous section. However, this time only right projection vectors  ${R^j}\in {\rm I\!R}^{32\times 1}|_{j=1}^2$ exist. Then, UMVCCA is run and the learned projection vectors are obtained. Fig. (\ref{mode1}) plots the true and learned projection vectors. It can be observed that without regarding the sign and scaling factor the true and learned vectors are similar.

\subsection{NIR-VIS 2.0 face dataset}
\begin{figure}
\includegraphics[scale=0.5]{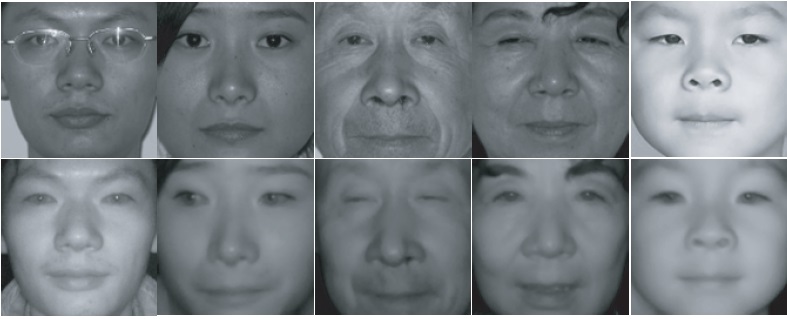}
\caption{Samples from NIR-VIS face dataset. First row (Visible images), Second row (Infrared images) }
\centering
\label{nir_vis_sam}
\end{figure}

We analyze the performance of the proposed algorithms on "NIR-VIS 2.0" \cite{li2013casia}.%  and "YaleB" \cite{GeBeKr01} face datasets.
This dataset contains the visible and near infrared face images. The data are collected through four different sessions and in each session some visible and infrared images are taken from each subject participated in that session. There are 740 unique "person-session" subjects in the dataset ( 710 persons were involved in only one session, while 15 persons were participated in two sessions). There are 1-22 VIS and 5-50 NIR face images per subject. However, some of the subjects only have one image and also some of them do not have either the visible or infrared images. Therefore, we select 728 "person-session" for our experiments. The same label is considered for the subjects who have involved in two sessions. We select 728 VIS and 728 NIR face images as the train data and 4333 VIS face images as the test data. The faced images are cropped so that the eyes of all images have the same coordinates. The face images are gray scaled and resized to $32\times32$. Fig.  \ref{nir_vis_sam} shows several samples from this dataset.

\subsubsection{Image reconstruction}
\begin{figure*}
\includegraphics[scale=0.3]{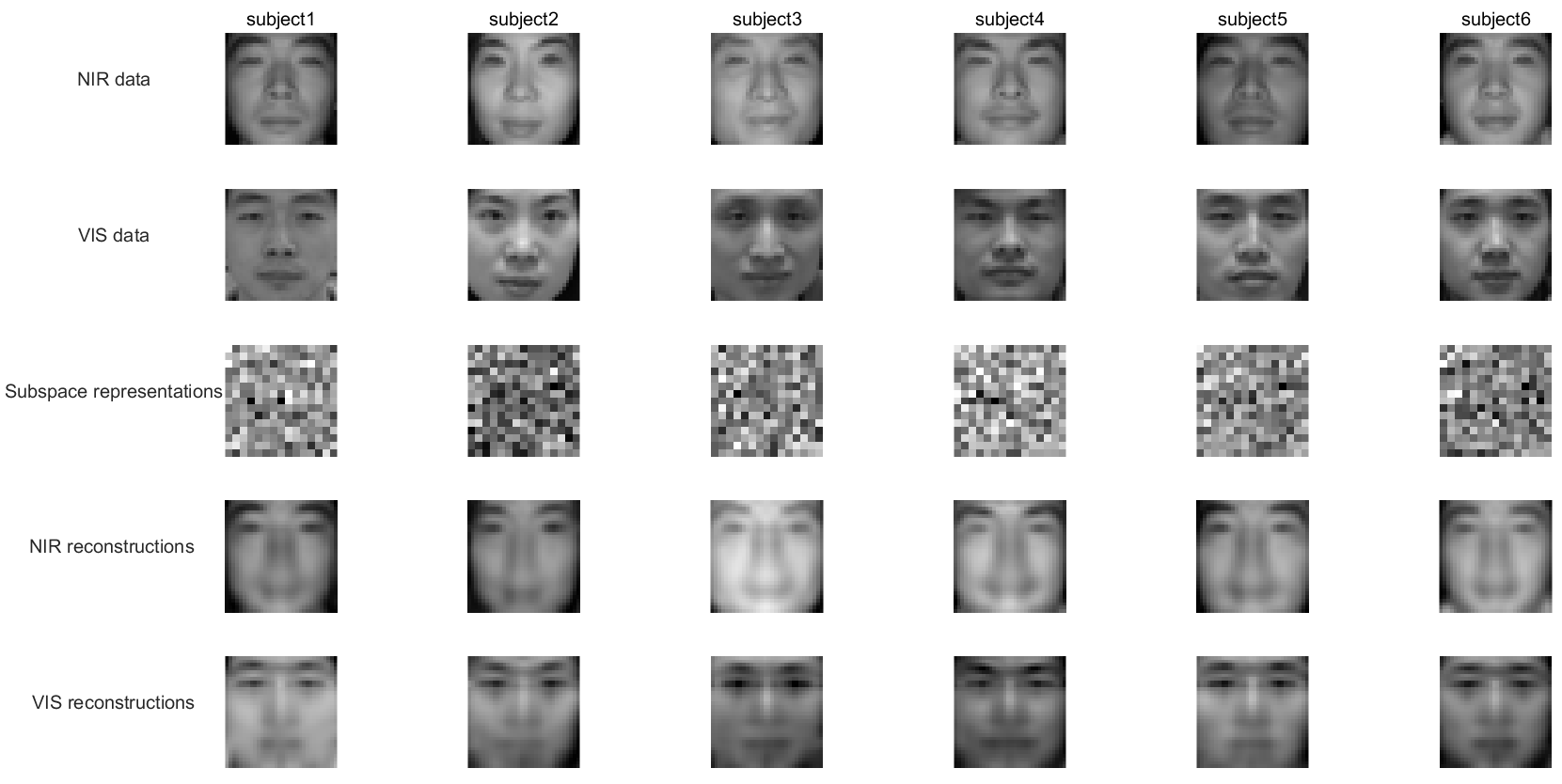}
\centering
\caption{Six persons original visible and infrared images and their corresponding reconstructed images.}
\label{sbspc_smpl}
\end{figure*}

In this experiment, we at first project the pairs of visible and infrared images into the low-dimensional subspace using BMVCCA and then try to reconstruct the original images. Fig. \ref{sbspc_smpl} depicts the images from visible and infrared spectrum of six persons as well as their corresponding low-dimensional subspace and reconstructed images. Here, compressed representation is $15\times 15$ and is obtained by equation (\ref{var5}). As it can be observed, reconstructed images are similar to the original images.

 \subsubsection{Convergence}
\begin{figure}[t!]
\begin{center}
%\begin{latin}
\input{"nir_vis_converge"}
%\end{latin}
\centering
\caption{Euclidean distance between consecutive projection matrices of BMVCCA in NIR-VIS 2.0 dataset}
\label{fig4:3}
\end{center}
\end{figure}
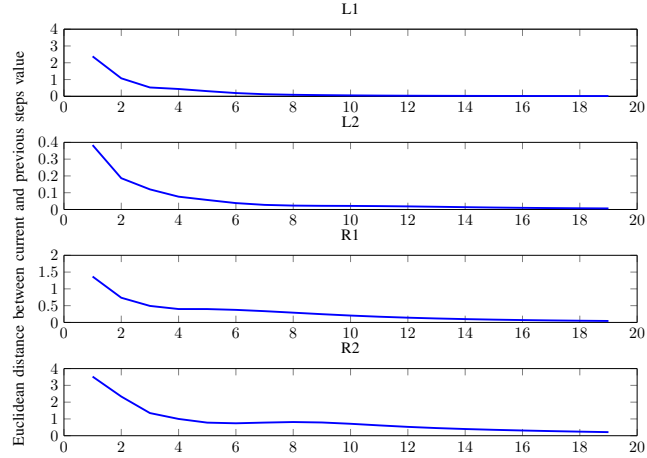

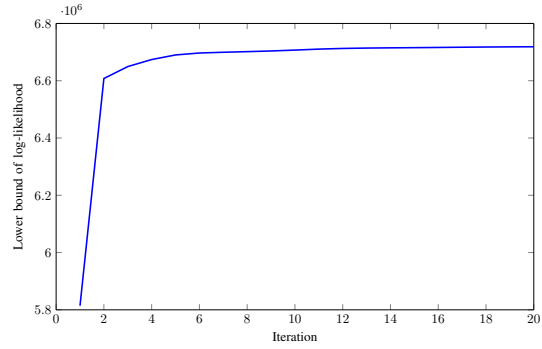
\begin{figure}
\centering
%\begin{latin}
\input{"fig_lower_bound"}
%\end{latin}
\caption{Lower bound of log-likelihood in BMVCCA algorithm in NIR-VIS 2.0 dataset}
\label{fig4:4}
\end{figure}

Fig. \ref{fig4:3} demonstrates the Euclidean distance of consecutive projection matrices in learning of BMVCCA. As it can be observed after some iterations the distance goes to zero. Also, Fig. \ref{fig4:4} depicts a plot of the lower bound of the logarithm of likelihood with respect to different iterations. As it can be observed the algorithm converges after some iterations.

\subsubsection{face recognition}
The task here is face recognition which we train the model with pairs of visible and infrared images and then project each train or test data into the low-dimensional space. Then compute the Euclidean distance between each projected test image and all the projected train images and choose the label of the train image with the least distance as the label of the test image. Table \ref{tab000} expresses the corresponding formula for the representing images into the low-dimensional space for each method. Moreover, in addition to the mentioned criteria based on the Euclidean distance between the projected test image and train images, we utilize the following probabilistic criteria: % we compute the label of test image $X$ using following formula:
 \begin{table}
% \begin{latin}
\begin{center}
    \begin{tabular}{ | l | c |}
    \hline
    Method & Subspace representation \\
    \hline
    CCA  & $ W^jx^j $ \\
    \hline
    PCCA\cite{bach2005probabilistic} & $ \mathbb{E}[z|x] $ \\
    \hline
    2DCCA\cite{lee2007two} & $ L^jX^jR^j $  \\
    \hline
    P2DCCA\cite{Safayani2017} & $ \mathbb{E}[Z|X] $  \\
    \hline
    UMVCCA & $ \mathbb{E}[Z|X] $ \\
    \hline
    BMVCCA & $  \mathbb{E}[q(Z)] $ \\
    \hline
    \end{tabular}
\end{center}
%\end{latin}
\caption{Different Methods and their low-dimensional representation formula}
\label{tab000}
 \end{table}

\begin{align} \label{var6}
&label(X^j)=\\\nonumber
&\argmax_n\mathbb{E}_q \Big[ ln[P(X^j|Z_n)]\Big], \quad {n=1...N},\quad j\in \{1,2\},
\end{align}
where here $X^j$ is a test image. This equation calculates the conditional probability of a test image given each of the training images and selects the label corresponding to the most probable image as the test data label. We call this approach probabilistic test and abbreviate it as "ptest" in the tables. Table \ref{nir_res} compares the error rate of different algorithms with different number of features in the low-dimensional space. For UMVCCA, the number of features is $32\times d$, where $d$ is the dimension of reduced feature space, therefore we select appropriate $d$ so that the number of obtained features is closed to the number of features in the corresponding column.  For example, the number of features in the first and second column of table \ref{nir_res} is $25$ and $100$ respectively. So for UMVCCA we consider the value of $d$ to be $1$ and $3$ respectively for these two columns, which produces $32$ and $96$ features. Since CCA and PCCA methods suffer from the small sample size problem that leads to the singular matrices. We at first project the data to the lower dimension of 727 (one less than the number of training data) using PCA and then applying the corresponding algorithms. Therefore, we cannot project the data to $30\times 30$ features for these algorithms and consequently place dash in the corresponding column. As it can be observed form table \ref{nir_res}, the BMVCCA with ptest criteria outperforms other algorithms significantly.

 \begingroup
\begin{table*}[t]
\renewcommand{\arraystretch}{1.3}
\caption{Comparison of error rate of different CCA based methods on NIR-VIS 2.0}
\centering
%\begin{latin}
\begin{tabular}{lccccccc}
\toprule
 & \multicolumn{6}{c}{Subspace dimentions} \\
\cmidrule(r){2-7}
Methods     & 5$\times$5 & 10$\times$10 & 15$\times$15 & 20$\times$20 & 25$\times$25 & 30$\times$30 & best   \\
\midrule
PCA+CCA     &  $89.2\%$  &   $51.4\%$   &   $31.1\%$   &    $  21.2$  &   $16.3\%$   & $---$     & $16.3\%$      \\
PCA+PCCA    &  $58.5\%$  &   $29.3\%$   &   $  20.7\%$   &    $16.6\%$  &   $15.4\%$   & $---$     & $  15.4\%$     \\
2DCCA       &  $27.2\%$  &   $  20.2\%$   &   $19.2\%$   &    $18.8\%$  &   $19.1\%$   & $19.3\%$     & $18.8\%$     \\
P2DCCA       &  $35.6\%$  &   $  18.7\%$   &   $11.3\%$   &    $11.1\%$  &   $9.2\%$   & $11.6\%$     & $9.2\%$     \\
UMVCCA      &  $22.9\%$  &   $15.6\%$   &   $25.1\%$   &   $29.6\%$   &   $  30\%$   & $  30\%$     & $  15.6\%$     \\
BMVCCA      &  $78.6\%$  &   $29.6\%$   &   $17.4\%$   &    $17.3\%$  &   $17.1\%$   & $17.6\%$     & $17.1\%$      \\
BMVCCA(ptest)&  $  40\%$  &   $13.2\%$   &   $ 9.4\%$   &   $9.1\%$   &   $7.5\%$    & $ 7.5\%$     & $ 7.5\%$       \\

\bottomrule
\end{tabular}
%\end{latin}
\label{nir_res}
\end{table*}
\endgroup

\section{Conclusion}\label{secconc}
In this paper a probabilistic model for CCA was proposed which works with matrix-variate data such as image matrices. Two iterative approach for learning the parameter were presented. In the first approach, called unilateral matrix variate CCA (UMVCCA), the model was restricted by mapping the latent matrix only from one side (row or column) and a learning method based on expectation maximization was introduced. In the other approach, called bilateral matrix variate CCA (BMVCCA), the latent matrix was mapped from the both sides and the posterior distribution was estimated using a variational matrix-variate distribution and its variational parameters were estimated using variational expectation maximization. The proposed algorithms was evaluated using syntectic and real data. The results indicated that the algorithm converges after a few iteration. Also, comparison with other CCA based algorithms showed that the proposed algorithm has a better performance in terms of recognition accuracy. This model can be extended to more complex models such as mixture models, nonlinear models, or bayesian framework.
\ifCLASSOPTIONcaptionsoff
  \newpage
\fi

% trigger a \newpage just before the given reference
% number - used to balance the columns on the last page
% adjust value as needed - may need to be readjusted if
% the document is modified later
%\IEEEtriggeratref{8}
% The "triggered" command can be changed if desired:
%\IEEEtriggercmd{\enlargethispage{-5in}}

% references section

% can use a bibliography generated by BibTeX as a .bbl file
% BibTeX documentation can be easily obtained at:
% http://mirror.ctan.org/biblio/bibtex/contrib/doc/
% The IEEEtran BibTeX style support page is at:
% http://www.michaelshell.org/tex/ieeetran/bibtex/
\bibliographystyle{IEEEtran}
% argument is your BibTeX string definitions and bibliography database(s)
\bibliography{biblio}
\end{document}

%% file: fig_synthetic_data_convergence.tex
% This file was created by matlab2tikz.
%
%The latest updates can be retrieved from
%  http://www.mathworks.com/matlabcentral/fileexchange/22022-matlab2tikz-matlab2tikz
%where you can also make suggestions and rate matlab2tikz.
%
\definecolor{mycolor1}{rgb}{0.00000,0.75000,0.75000}%
\begin{tikzpicture}[scale=0.8]
\begin{axis}[%
width=8cm,
height=5cm,
at={(2.152in,0.893in)},
scale only axis,
separate axis lines,
every outer x axis line/.append style={black},
every x tick label/.append style={font=\color{black}},
xmin=1,
xmax=10,
xlabel={Iteration},
every outer y axis line/.append style={black},
every y tick label/.append style={font=\color{black}},
ymin=0,
ymax=18,
ylabel={Distance},
axis background/.style={fill=white},
legend style={legend cell align=left,align=left,draw=black}
]
\addplot [color=black!20!brown,solid, mark=triangle]
  table[row sep=crcr]{%
1	7.39705580639686\\
2	2.024003455428\\
3	2.36221523144148\\
4	3.77557002213725\\
5	4.81822137806479\\
6	4.60464408131686\\
7	0.608029020851496\\
8	0.131215183982326\\
9	0.0356885406450998\\
10	0.00996234070574813\\
};
\addlegendentry{$L^1$};

\addplot [color=green!50!black,solid,mark=triangle*]
  table[row sep=crcr]{%
1	8.00673612374311\\
2	2.45973658659297\\
3	2.7509046176294\\
4	4.37976249032033\\
5	7.22063444107824\\
6	6.67514489128337\\
7	0.728502212810631\\
8	0.152064410029832\\
9	0.0395997597472932\\
10	0.0108956831986857\\
};
\addlegendentry{$L^2$};

\addplot [color=blue,solid,mark=o]
  table[row sep=crcr]{%
1	17.6841402539971\\
2	4.43845825415012\\
3	2.71306932492719\\
4	3.05220489474997\\
5	3.59858412221797\\
6	0.595600422614447\\
7	0.143295153136647\\
8	0.0256402610016238\\
9	0.00694920550336545\\
10	0.00192559037025702\\
};
\addlegendentry{$R^1$};

\addplot [color=red,solid,mark=*]
  table[row sep=crcr]{%
1	11.4774163339339\\
2	3.22062026043411\\
3	2.97510762602607\\
4	5.66086914262899\\
5	5.48587147828988\\
6	2.07761625844707\\
7	0.12861870694238\\
8	0.0214552814246254\\
9	0.00579432741653564\\
10	0.00159699051679907\\
};
\addlegendentry{$R^2$};

\addplot [line width=0.3mm, color=black,solid,mark=square*]
  table[row sep=crcr]{%
1	11.1413371295177\\
2	3.0357046391513\\
3	2.70032420000603\\
4	4.21710163745914\\
5	5.28082785491272\\
6	3.48825141341544\\
7	0.402111273435288\\
8	0.0825937841096018\\
9	0.0220079583280735\\
10	0.00609515119787248\\
};
\addlegendentry{average};

\end{axis}
\end{tikzpicture}%

%% file: fig_dist2.tex
% This file was created by matlab2tikz.
%
%The latest updates can be retrieved from
%  http://www.mathworks.com/matlabcentral/fileexchange/22022-matlab2tikz-matlab2tikz
%where you can also make suggestions and rate matlab2tikz.
%
\begin{tikzpicture}[scale=0.8]

\begin{axis}[%
width=8cm,
height=5cm,
at={(0.753in,0.478in)},
scale only axis,
separate axis lines,
every outer x axis line/.append style={black},
every x tick label/.append style={font=\color{black}},
xmin=1,
xmax=10,
xlabel={Iteration},
every outer y axis line/.append style={black},
every y tick label/.append style={font=\color{black}},
ymin=0.0007,
ymax=0.0015,
ylabel={Distance},
axis background/.style={fill=white}
]
\addplot [color=blue,solid,forget plot,thick]
  table[row sep=crcr]{%
1	0.00135299504224854\\
2	0.00102843416365315\\
3	0.000921782923118346\\
4	0.000835845122408069\\
5	0.000718304574624369\\
6	0.000713962419774156\\
7	0.000712739092344934\\
8	0.000713507279612859\\
9	0.000714293753147567\\
10	0.000715108771838799\\
};
\end{axis}
\end{tikzpicture}% 

%% file: fig_dist_to_true_post.tex
% This file was created by matlab2tikz.
%
%The latest updates can be retrieved from
%  http://www.mathworks.com/matlabcentral/fileexchange/22022-matlab2tikz-matlab2tikz
%where you can also make suggestions and rate matlab2tikz.
%
\begin{tikzpicture}[scale=0.8]

\begin{axis}[%
compat=newest,
width=8cm,
height=5cm,
at={(0.753in,0.478in)},
scale only axis,
separate axis lines,
every outer x axis line/.append style={black},
every x tick label/.append style={font=\color{black}},
xmode=log,
xmin=10,
xmax=1000,
xminorticks=true,
xlabel={Number of training samples},
every outer y axis line/.append style={black},
every y tick label/.append style={font=\color{black}},
ymin=0,
ymax=0.14,
ylabel={Distance},
axis background/.style={fill=white}
]
\addplot [color=blue,solid,forget plot,thick]
  table[row sep=crcr]{%
10	0.130032040213438\\
15	0.082329531248349\\
21	0.0670444074159152\\
28	0.0472054966472258\\
36	0.0375150657957134\\
44	0.0289284973495117\\
53	0.0252920812642778\\
62	0.0214943987282662\\
72	0.0183659527058414\\
82	0.0168594349362014\\
93	0.0147064192789828\\
104	0.0130889061488873\\
115	0.0120761875011749\\
126	0.010918440155281\\
138	0.0101206175443044\\
150	0.00912527042217688\\
162	0.00841908587778582\\
174	0.00780596199779971\\
186	0.00730531165949106\\
199	0.00644143412237793\\
212	0.00657412084218001\\
225	0.00610452257343165\\
238	0.00577417446060474\\
251	0.00552245861199135\\
264	0.0046008356641652\\
277	0.00505739669223894\\
291	0.00472155467183976\\
305	0.00447889152600693\\
319	0.00417297368021553\\
333	0.00411288051832538\\
347	0.00403106115211803\\
361	0.00357990000173968\\
375	0.00360788201838415\\
389	0.0035653071371293\\
403	0.00345057873065302\\
417	0.00340504456956616\\
432	0.00318429516865839\\
447	0.00313112366370097\\
462	0.00288334021993045\\
477	0.00279190891162643\\
492	0.00277958685085071\\
507	0.00273233107759431\\
522	0.00266022584105917\\
537	0.00260396086856584\\
552	0.00253550062842629\\
567	0.00248369052723411\\
582	0.00236771529842269\\
597	0.0021763033433455\\
612	0.00227732844831144\\
628	0.00221792900894633\\
644	0.00208307112132477\\
660	0.00207466952520686\\
676	0.00208066163830811\\
692	0.00201623506602167\\
708	0.00194060532531758\\
724	0.00179999270735843\\
740	0.00188652670734974\\
756	0.00177497575889407\\
772	0.00172944032624766\\
788	0.00177637089389557\\
804	0.00172800733324068\\
820	0.00167701249054203\\
836	0.00153537911177511\\
852	0.00164991077996286\\
868	0.00153668584476126\\
884	0.00159784774253193\\
900	0.0014890958525518\\
917	0.00152592397854714\\
934	0.0014951472208499\\
951	0.00145228657599883\\
968	0.00128964793453497\\
985	0.00141272603467359\\
};
\end{axis}
\end{tikzpicture}% 

%% file: myfile.tex
% This file was created by matlab2tikz.
%
%The latest updates can be retrieved from
%  http://www.mathworks.com/matlabcentral/fileexchange/22022-matlab2tikz-matlab2tikz
%where you can also make suggestions and rate matlab2tikz.
%
\begin{tikzpicture}[scale=0.65]

\begin{axis}[%
width=2.172in,
height=1.692in,
at={(0.844in,2.895in)},
ylabel={$R^1$},
y label style={at={(-0.1,0.5)}},
ylabel style={rotate=-90},
scale only axis,
xmin=1,
xmax=32,
ymin=-4,
ymax=4,
axis background/.style={fill=white},
]
\addplot [color=blue,solid,mark=square,mark options={solid},forget plot]
  table[row sep=crcr]{%
1	-1.0700609192233\\
2	-0.218814318738795\\
3	-0.451068139477732\\
4	-1.35359240989006\\
5	-0.779669021908806\\
6	0.364369134508533\\
7	-0.765191541436278\\
8	-0.734537298763098\\
9	3.17481623319339\\
10	-0.659769167818569\\
11	-1.89579644645245\\
12	-1.401734996825\\
13	-1.12910014375132\\
14	-1.35433454494928\\
15	0.921171867040565\\
16	1.47059071118788\\
17	-0.651979853296216\\
18	-0.507591815171417\\
19	2.10293189242209\\
20	-2.51677706543649\\
21	0.411367192740334\\
22	1.20144854015312\\
23	-1.00870126937729\\
24	-0.585142698403563\\
25	0.256328643065044\\
26	0.419150260949317\\
27	0.356832262520743\\
28	1.38528976477017\\
29	-0.856521558204858\\
30	-0.191572773072754\\
31	0.476171924212795\\
32	-0.415105651586245\\
};
\end{axis}

\begin{axis}[%
width=2.172in,
height=1.692in,
at={(3.701in,2.895in)},
ylabel={$\tilde{R}^1$},
y label style={at={(-0.1,0.5)}},
ylabel style={rotate=-90},
scale only axis,
xmin=1,
xmax=32,
ymin=-4,
ymax=6,
axis background/.style={fill=white},
]
\addplot [color=red,solid,mark=square,mark options={solid},forget plot]
  table[row sep=crcr]{%
1	-1.59334219170748\\
2	-0.368879145294364\\
3	-0.712886134081464\\
4	-2.09570025402115\\
5	-1.12007837071317\\
6	0.568881368480287\\
7	-1.1554896892109\\
8	-1.1291011103802\\
9	4.86809754987803\\
10	-1.00517481103134\\
11	-2.86132266661313\\
12	-2.09445481652967\\
13	-1.68634798836478\\
14	-2.08884383824507\\
15	1.3784672445669\\
16	2.23403409361194\\
17	-1.02244643595671\\
18	-0.746240665720981\\
19	3.21476493999534\\
20	-3.75501120716694\\
21	0.646224601166847\\
22	1.84294628116694\\
23	-1.54758366138965\\
24	-0.909913452476499\\
25	0.44133567808232\\
26	0.695134034399366\\
27	0.551919424518191\\
28	2.06303022129433\\
29	-1.30578166236274\\
30	-0.291894718793028\\
31	0.718656945054931\\
32	-0.64273084655092\\
};
\end{axis}

\begin{axis}[%
width=2.172in,
height=1.692in,
at={(0.844in,0.545in)},
ylabel={$R^2$},
y label style={at={(-0.1,0.5)}},
ylabel style={rotate=-90},
scale only axis,
xmin=1,
xmax=32,
xlabel={(a)},
ymin=-3,
ymax=3,
axis background/.style={fill=white}
]
\addplot [color=blue,solid,mark=o,mark options={solid},forget plot]
  table[row sep=crcr]{%
1	0.125540964205318\\
2	-2.54092318887262\\
3	-0.317319089413717\\
4	-0.0181429135940753\\
5	-0.00536663973497835\\
6	-1.14863179715817\\
7	-0.526768710279969\\
8	0.317022119843051\\
9	0.506676405498338\\
10	-1.64724316045225\\
11	-0.848325314693811\\
12	1.37815402713207\\
13	0.667190541472069\\
14	-0.19964922032369\\
15	0.219840843349621\\
16	0.994251040494133\\
17	-0.776522643346993\\
18	0.163505960361634\\
19	-1.0605492211971\\
20	0.0590855544066489\\
21	-1.44254046163233\\
22	0.415909579888521\\
23	-0.624041613582736\\
24	0.494677089460974\\
25	-1.50340723331645\\
26	-0.214803539458743\\
27	2.11344996512879\\
28	-1.15233098861098\\
29	-0.287256846475346\\
30	1.27603967077757\\
31	1.54742819626119\\
32	1.17489826994265\\
};
\end{axis}

\begin{axis}[%
width=2.172in,
height=1.692in,
at={(3.701in,0.545in)},
ylabel={$\tilde{R}^2$},
y label style={at={(-0.1,0.5)}},
ylabel style={rotate=-90},
scale only axis,
xmin=1,
xmax=32,
xlabel={(b)},
ymin=-4,
ymax=4,
axis background/.style={fill=white}
]
\addplot [color=red,solid,mark=o,mark options={solid},forget plot]
  table[row sep=crcr]{%
1	0.222181793499709\\
2	-3.82546408568591\\
3	-0.49163531402921\\
4	0.0130500371751253\\
5	-0.0279651198242983\\
6	-1.8056445798228\\
7	-0.786820243331301\\
8	0.48421681417956\\
9	0.802666382567907\\
10	-2.48876603591911\\
11	-1.24668672353662\\
12	2.07117604019703\\
13	0.988083034510612\\
14	-0.289367306283641\\
15	0.347508314035899\\
16	1.48705641451084\\
17	-1.16871069313358\\
18	0.311980066984041\\
19	-1.59922153801543\\
20	0.0795382581410238\\
21	-2.18672843449937\\
22	0.652473255181431\\
23	-0.941014626248081\\
24	0.781097488658046\\
25	-2.26240623607657\\
26	-0.357207816799839\\
27	3.20460665640373\\
28	-1.72557250593485\\
29	-0.434542849458729\\
30	1.8936038186321\\
31	2.38652506839462\\
32	1.7870813969258\\
};
\end{axis}
\end{tikzpicture}%

%% file: nir_vis_converge.tex
% This file was created by matlab2tikz.
%
%The latest updates can be retrieved from
%  http://www.mathworks.com/matlabcentral/fileexchange/22022-matlab2tikz-matlab2tikz
%where you can also make suggestions and rate matlab2tikz.
%
\begin{tikzpicture}[scale=0.6]

\begin{axis}[thick,%
width=5in,
height=0.586in,
at={(1.325in,3.459in)},
scale only axis,
separate axis lines,
every outer x axis line/.append style={black},
every x tick label/.append style={font=\color{black}},
xmin=0,
xmax=20,
every outer y axis line/.append style={black},
every y tick label/.append style={font=\color{black},thick},
ymin=0,
ymax=4,
axis background/.style={fill=white},
title={L1}
]
\addplot [color=blue,solid,forget plot,very thick]
  table[row sep=crcr]{%
1	2.37995347957154\\
2	1.07268800261812\\
3	0.526949807003006\\
4	0.436641032171541\\
5	0.308083591943006\\
6	0.19397041717182\\
7	0.124403584726115\\
8	0.0900742021080041\\
9	0.0704851464582668\\
10	0.0550326520609743\\
11	0.0435808826113147\\
12	0.0354895311272967\\
13	0.0297764687519828\\
14	0.0255479509162886\\
15	0.0222551039842476\\
16	0.0196241417659938\\
17	0.0175221485869942\\
18	0.0158757777238812\\
19	0.0146322980601992\\
};
\end{axis}

\begin{axis}[%
width=5in,
height=0.586in,
at={(1.325in,2.471in)},
scale only axis,
separate axis lines,
every outer x axis line/.append style={black},
every x tick label/.append style={font=\color{black}},
xmin=0,
xmax=20,
every outer y axis line/.append style={black},
every y tick label/.append style={font=\color{black}},
ymin=0,
ymax=0.4,
axis background/.style={fill=white},
title={L2}
]
\addplot [color=blue,solid,forget plot,very thick]
  table[row sep=crcr]{%
1	0.383777135752126\\
2	0.186499407390077\\
3	0.119819346350058\\
4	0.0763348592746993\\
5	0.0567877717455788\\
6	0.0380913135322389\\
7	0.0274544975316599\\
8	0.023177054866662\\
9	0.0218218252313667\\
10	0.0212763809999531\\
11	0.0201971097886155\\
12	0.0184822644653158\\
13	0.0162440020131089\\
14	0.0138237085449578\\
15	0.0116326204724816\\
16	0.00982618736554527\\
17	0.0083818618723533\\
18	0.0072300958242047\\
19	0.00630696670097362\\
};
\end{axis}

\begin{axis}[%
width=5in,
height=0.586in,
at={(1.325in,1.485in)},
scale only axis,
separate axis lines,
every outer x axis line/.append style={black},
every x tick label/.append style={font=\color{black}},
xmin=0,
xmax=20,
every outer y axis line/.append style={black},
every y tick label/.append style={font=\color{black}},
ymin=0,
ymax=2,
axis background/.style={fill=white},
title={R1},
ylabel={Euclidean distance between current and previous steps value},every axis y label/.append style={at=(ticklabel cs:1.1)}
]
\addplot [color=blue,solid,forget plot,very thick]
  table[row sep=crcr]{%
1	1.36727636562309\\
2	0.735673881280332\\
3	0.491904972258151\\
4	0.398896084409986\\
5	0.398145262276577\\
6	0.374082023480713\\
7	0.335919672328005\\
8	0.291082712208481\\
9	0.245901039207495\\
10	0.20499311353157\\
11	0.170160620152838\\
12	0.141443163910376\\
13	0.118049899277791\\
14	0.0989603809438459\\
15	0.0832991239003919\\
16	0.0704094541491435\\
17	0.0597669321483271\\
18	0.0509489862390171\\
19	0.0436362141376124\\
};
\end{axis}

\begin{axis}[%
width=5in,
height=0.586in,
at={(1.325in,0.496in)},
scale only axis,
separate axis lines,
every outer x axis line/.append style={black},
every x tick label/.append style={font=\color{black}},
xmin=0,
xmax=20,
every outer y axis line/.append style={black},
every y tick label/.append style={font=\color{black}},
ymin=0,
ymax=4,
axis background/.style={fill=white},
title={R2}
]
\addplot [color=blue,solid,forget plot,very thick]
  table[row sep=crcr]{%
1	3.52233071265774\\
2	2.33458330838179\\
3	1.3495607131958\\
4	0.999523804099934\\
5	0.779277656206288\\
6	0.740670321887708\\
7	0.782300545997213\\
8	0.815130085538539\\
9	0.788742143627645\\
10	0.711062799366783\\
11	0.615193464908365\\
12	0.526940473687688\\
13	0.455480899937276\\
14	0.398782227426048\\
15	0.351746640675295\\
16	0.310975045685983\\
17	0.274861073489389\\
18	0.242563880942968\\
19	0.213572605701828\\
};
\end{axis}
\end{tikzpicture}%

%% file: fig_lower_bound.tex
% This file was created by matlab2tikz.
%
%The latest updates can be retrieved from
%  http://www.mathworks.com/matlabcentral/fileexchange/22022-matlab2tikz-matlab2tikz
%where you can also make suggestions and rate matlab2tikz.
%
\begin{tikzpicture}[scale=0.5]

\begin{axis}[%
width=5in,
height=3in,
at={(2.152in,0.893in)},
scale only axis,
separate axis lines,
every outer x axis line/.append style={black},
every x tick label/.append style={font=\color{black}},
xmin=0,
xmax=20,
every outer y axis line/.append style={black},
every y tick label/.append style={font=\color{black}},
ymin=5800000,
ymax=6800000,
axis background/.style={fill=white},
ylabel={Lower bound of log-likelihood},
xlabel={Iteration}
]
\addplot [color=blue,solid,forget plot,very thick]
  table[row sep=crcr]{%
1	5814264.59081461\\
2	6607724.12768094\\
3	6649421.1303697\\
4	6673968.05259326\\
5	6690008.09580698\\
6	6696800.26234044\\
7	6699402.19842844\\
8	6701543.99651601\\
9	6703887.7365898\\
10	6707009.2535282\\
11	6710393.3795268\\
12	6712726.98790511\\
13	6714033.3973972\\
14	6714835.75411652\\
15	6715474.48674717\\
16	6716154.93453259\\
17	6716973.37339644\\
18	6717725.54824289\\
19	6718229.12598051\\
20	6718611.3533461\\
};
\end{axis}
\end{tikzpicture}%